\documentclass[runningheads]{llncs}
\usepackage{pifont}
\usepackage[T1]{fontenc}
\usepackage{graphicx}
\usepackage{amssymb}
\usepackage{amsmath}
\usepackage[misc]{ifsym}
\usepackage{color, xcolor}
\usepackage{multirow}
\usepackage{multicol}
\usepackage{colortbl}
\usepackage{bbding}
\usepackage{float}
\usepackage{bm}
\usepackage{soul}
\usepackage{tabularx,booktabs}
\usepackage{hyperref}
\usepackage[most]{tcolorbox}
\hypersetup{colorlinks=true, linkcolor=blue, anchorcolor=blue, citecolor=blue}

\begin{document}

\title{CAS-IQA: Teaching Vision-Language Models for Synthetic Angiography Quality Assessment}
\titlerunning{CAS-IQA}

\author{
Bo Wang\inst{1,3}$^\star$ \and
De-Xing Huang\inst{1,2}\thanks{Equal contributions.} \and
Xiao-Hu Zhou\inst{1,2}\textsuperscript{(\Letter)} \and
Mei-Jiang Gui\inst{1,2} \and
Nu-Fang Xiao\inst{1,4} \and
Jian-Long Hao\inst{5} \and
Ming-Yuan Liu\inst{6}\textsuperscript{(\Letter)} \and
Zeng-Guang Hou\inst{1,2}
}

\authorrunning{B. Wang et al.}

\institute{
State Key Laboratory of Multimodal Artificial Intelligence Systems, \\ Institute of Automation, Chinese Academy of Sciences
\and
School of Artificial Intelligence, University of Chinese Academy of Sciences
\and
School of Automation and Electrical Engineering, \\ University of Science and Technology Beijing
\and
School of Computer Science and Engineering, \\ Hunan University of Science and Technology
\and
School of Information, Shanxi University of Finance and Economics
\and
Beijing Friendship Hospital, Capital Medical University
\\
Email: \email{\href{mailto:xiaohu.zhou@ia.ac.cn}{xiaohu.zhou@ia.ac.cn}, \href{mailto:dr.mingyuanliu@pku.edu.cn}{dr.mingyuanliu@pku.edu.cn}}
}

\maketitle              

\begin{abstract}
Synthetic X-ray angiographies generated by modern generative models hold great potential to reduce the use of contrast agents in vascular interventional procedures. However, low-quality synthetic angiographies can significantly increase procedural risk, underscoring the need for reliable image quality assessment (IQA) methods. Existing IQA models, however, fail to leverage auxiliary images as references during evaluation and lack fine-grained, task-specific metrics necessary for clinical relevance. To address these limitations, this paper proposes CAS-IQA, a vision-language model (VLM)-based framework that predicts fine-grained quality scores by effectively incorporating auxiliary information from related images. In the absence of angiography datasets, CAS-3K is constructed, comprising $3,565$ synthetic angiographies along with score annotations. To ensure clinically meaningful assessment, three task-specific evaluation metrics are defined. Furthermore, a \underline{M}ulti-path feat\underline{U}re fu\underline{S}ion and rou\underline{T}ing (MUST) module is designed to enhance image representations by adaptively fusing and routing visual tokens to metric-specific branches. Extensive experiments on the CAS-3K dataset demonstrate that CAS-IQA significantly outperforms state-of-the-art IQA methods by a considerable margin.

\end{abstract}
\section{Introduction} \label{sec:introduction}
\begin{figure}
\centering
\centerline{\includegraphics{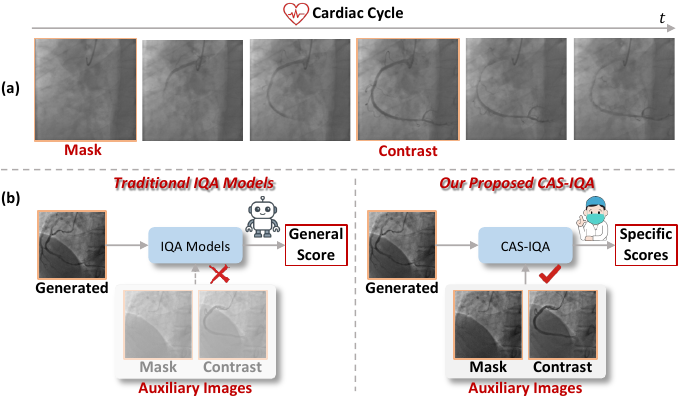}}
\caption{(a) An X-ray sequence in real-world clinical scenarios. (b) Comparison between current IQA models and our CAS-IQA.}
\label{fig4}
\end{figure}

X-ray angiography is regarded as the ``gold standard'' for diagnosing and treating cardiovascular diseases~\cite{zhao2024large}. However, contrast agents used in these procedures pose significant health risks to patients, including allergic reactions~\cite{clement2018immediate} and renal dysfunction~\cite{fahling2017understanding}. Recent advances in AI-generated content (AIGC)~\cite{rombach2022high} have demonstrated remarkable capabilities in generating photo‑realistic angiographies from non‑contrast X‑ray images~\cite{huang2024cas}, presenting a promising way to reduce reliance on contrast agents in clinical scenarios. Nevertheless, \textit{the fidelity of such synthetic angiographies to essential clinical constraints remains unquantified}. This highlights the urgent need for rigorous quality assessment methods.

Image quality assessment (IQA) is a fundamental task in computer vision~\cite{wu2024qalign}. It aims to bridge quantitative mappings between images and corresponding human subjective judgments~\cite{mittal2012no}. Modern IQA methods can achieve strong quality assessment performance on natural images~\cite{su2020blindly,xu2024boosting,wu2024qalign,you2025teaching}, but they falter in our specialized domain for two reasons: \textbf{1) Inability to Leverage Auxiliary Images.} Fig.~\ref{fig4} (a) presents a real X-ray sequence in clinical practice. The first image without contrast agents is defined as \textbf{Mask}, which is used to generate angiographies (\textbf{Generated}) by generative models. The image full with contrast agents is called \textbf{Contrast}\footnote{Note that \textbf{Mask} and \textbf{Generated} are captured at different moments in the cardiac cycle. The vessels may have undergone slight deformation due to the heartbeat.}. Obviously, it is essential to incorporate both \textbf{Mask} and \textbf{Contrast} when assessing the quality of \textbf{Generated}. However, current IQA models typically operate on \textbf{Generated} and lack mechanisms to effectively fuse multi-input information; \textbf{2) Lack of Fine-grained, Task-specific Metrics.} Current IQA methods often produce a single, abstract quality score, which is insufficient for clinical applications. In practice, synthetic angiographies are primarily used to visualize vascular structures, and radiologists are particularly concerned with the structural integrity and anatomical accuracy of these vessels. Therefore, a more nuanced assessment framework is needed.

To address these drawbacks, \textbf{CAS-IQA} (\textbf{C}ontrast-free \textbf{A}ngiography \textbf{S}ynthesis \textbf{I}mage \textbf{Q}uality \textbf{A}ssessment) is proposed, a novel framework for evaluating the quality of synthetic angiographies. Given the lack of available datasets, CAS-3K is constructed, a high-quality benchmark dataset comprising over $3,500$ synthetic angiographies generated by five state-of-the-art generative models. To comprehensively evaluate synthetic angiographies, three metrics verified by board-certified radiologists are introduced: \textbf{1) Vessel Morphology Consistency (VMC)} assesses the structural consistency between \textbf{Generated} and \textbf{Contrast}; \textbf{2) Vessel Branch Detection (VBD)} evaluates whether key vascular branches are correctly generated in \textbf{Generated}; \textbf{3) Overall Quality (OQ)} considers not only the first two metrics, but also visual defects such as background artifacts or shadowing in \textbf{Generated}.

Based on the CAS-3K dataset, CAS-IQA is constructed. Specifically, CAS-IQA is built upon vision-language models (VLMs), as VLMs allow flexible evaluation of different aspects of images through prompt engineering, better aligning with human subjective judgments. Moreover, a \underline{M}ulti-path feat\underline{U}re fu\underline{S}ion and rou\underline{T}ing (MUST) module is presented, which fuses features of \textbf{Mask}, \textbf{Contrast}, and \textbf{Generated}, and routes them to metric-specific branches for optimized assessment.

Our main contributions are summarized as follows:
\begin{itemize}
    \item[$\bullet$] The first high-quality dataset for angiography IQA (CAS-3K) is constructed, consisting of $3,565$ synthetic angiographies with metric-based annotations.
    \item[$\bullet$] A novel IQA framework based on vision-language models is proposed to comprehensively evaluate the quality of synthetic angiographies. Moreover, the MUST module is designed to effectively fuse visual cues from multiple image sources and adapt them to different evaluation metrics.
    \item[$\bullet$] Extensive experimental results indicate that CAS-IQA significantly outperforms state-of-the-art IQA models on the CAS-3K dataset.
\end{itemize}

\section{Related Works} \label{sec:related_works}
\subsection{Image Quality Assessment}
As a fundamental task in computer vision, image quality assessment (IQA) has undergone substantial advancements across diverse applications. Early methods primarily rely on human-crafted metrics such as NIQE~\cite{mittal2012making} and BRISQUE~\cite{mittal2012no}, \textit{etc}. However, these approaches heavily depend on prior knowledge and often exhibit limited generalization capabilities~\cite{you2025teaching}. In recent years, the rapid development of deep neural networks (DNNs) has introduced robust solutions for IQA~\cite{su2020blindly,xu2024boosting,yang2022maniqa}. Typically formulated as a regression problem, DNN-based IQA methods extract visual features from input images using backbone encoders, and then predict corresponding quality scores. Thanks to strong feature extraction capabilities of advanced network architectures, these models have achieved impressive performance on standard IQA benchmarks. Nonetheless, they struggle to capture fine-grained semantic content of images, which is crucial for quality assessment~\cite{wang2024large}. To address this limitation, recent studies have explored the use of vision-language models (VLMs) for IQA~\cite{wu2024q,wu2024qalign}. By leveraging the advanced multimodal reasoning capabilities of VLMs, these methods have demonstrated substantial progress in aligning quality predictions with human subjective judgments. However, as discussed in Section~\ref{sec:introduction}, existing approaches fall short in our specific domain.

\subsection{Vision-Language Models}
Vision-language models (VLMs) extend large language models (LLMs) by incorporating visual instruction tuning~\cite{liu2023visual}, which enables them to jointly process and align visual and textual modalities. This alignment facilitates the development of visual reasoning and understanding capabilities, allowing VLMs to follow multimodal instructions and perform complex vision-language tasks in a unified framework (\textit{e.g.,} LLaVA~\cite{liu2023visual} and mPLUG-Owl2~\cite{ye2024mplug}), including visual question answering (VQA)~\cite{liu2024mmbench} and image captioning~\cite{hua2025finecaption},~\textit{etc}. In this paper, the proposed CAS-IQA builds upon the powerful visual reasoning capabilities of VLMs to achieve more accurate and human-aligned IQA.

\subsection{Datasets for Image Quality Assessment}
High-quality IQA datasets are the foundation for developing and benchmarking effective IQA models. Considerable efforts have been devoted to constructing diverse IQA datasets spanning both natural and AIGC. Traditional IQA datasets, such as KonIQ-10K~\cite{hosu2020koniq} and SPAQ~\cite{fang2020perceptual}, primarily focus on natural images captured under a wide range of real-world conditions, offering a broad spectrum of authentic distortions and degradation types. With the rapid emergence of advanced generative models, there has been a significant rise in AIGC, leading to the need for specialized datasets and methodologies to assess the perceptual quality of synthetic images. In response, datasets such as AGIQA-3K~\cite{li2023agiqa} and AIGIQA-20K~\cite{li2024aigiqa} have been proposed. To the best of our knowledge, CAS-3K is the first IQA dataset for medical AIGC.

\section{CAS-3K Dataset} \label{sec:cas3k}
\subsection{Image Collection and Scoring Rules}~\label{sec:cas3k_rules}
\begin{figure}
\centering
\centerline{\includegraphics{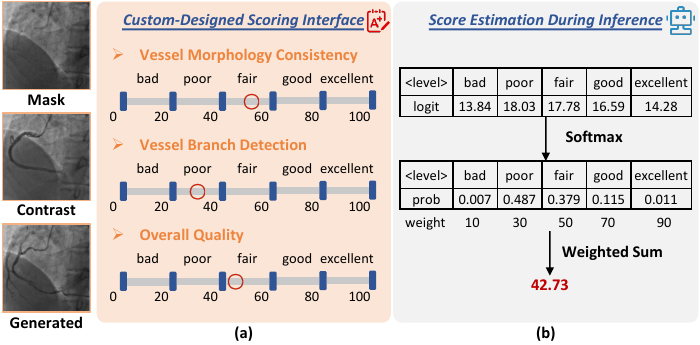}}
\caption{An overview of the rating GUI of CAS-3K and sore estimation rules during inference. (a) illustrates how five-level scores are assigned to rate the three quality metrics of \textbf{Generated} while using \textbf{Mask} and \textbf{Contrast} as references. (b) demonstrates how the model-predicted level logits are converted into final quality scores.}
\label{fig1}
\end{figure}
\textbf{Image Collection.} To construct CAS-3K, we curate high-quality synthetic angiographies using two publicly available datasets: XCAD~\cite{ma2021self} and CADICA~\cite{jimenez2024cadica}. After thorough manual screening, $713$ high-quality \textbf{Mask}-\textbf{Contrast} image pairs are selected from the two datasets. Using \textbf{Mask} as input, we apply five state-of-the-art generative models\footnote{Three GAN-based models (\textit{i.e.,} CycleGAN~\cite{zhu2017unpaired}, AttentionGAN~\cite{tang2021attentiongan}, and CUT~\cite{park2020contrastive}) and two diffusion-based models (\textit{i.e.,} ILVR~\cite{choi2021ilvr} and EGSDE~\cite{zhao2022egsde}).} to synthesize angiographies (\textbf{Generated}). This resulted in a total of $3,565$ \textbf{Generated}, each paired with its corresponding \textbf{Mask} and \textbf{Contrast}.

\noindent\textbf{Scoring Rules.} To ensure clinically relevant quality assessment, we collaborat with two board-certified radiologists to define three task-specific evaluation metrics:
\begin{itemize}
    \item [\textbf{1)}] \textbf{Vessel Morphology Consistency (VMC):} It assesses the anatomical accuracy of major vessels in \textbf{Generated} compared to \textbf{Contrast}. Key factors include the morphological fidelity of primary vessel structures and their continuity—specifically the absence of interruptions, or breaks.
    \item [\textbf{2)}] \textbf{Vessel Branch Detection (VBD):} It assesses the correctness of generated vascular branches. This includes evaluating the number of branches as well as the accuracy of their orientations and trajectories, as compared to those in \textbf{Contrast}.
    \item [\textbf{3)}] \textbf{Overall Quality (OQ):} This metric provides a holistic assessment by balancing VMC and VBD, while also accounting for visual artifacts such as background noise, shadowing, or unrealistic textures. This metric reflects the overall anatomical plausibility and visual realism of \textbf{Generated}.
\end{itemize}

To support fine-grained evaluation, a five-level rating scheme is adopted inspired by the Q-Align~\cite{wu2024qalign}. Each image is scored on a continuous $0\sim100$ scale, partitioned into five semantic categories: bad, poor, fair, good, and excellent.

\subsection{Subjective Quality Assessment}
\begin{figure}
\centering
\centerline{\includegraphics{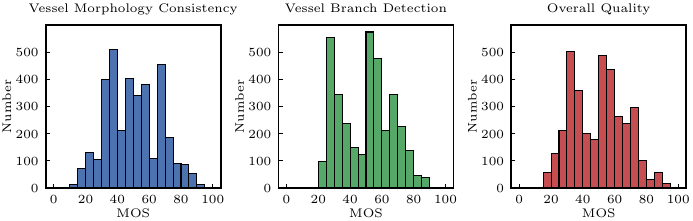}}
\caption{Distributions of mean opinion score (MOS).}
\label{fig_mos}
\end{figure}
Following ITU-R BT.$500$~\cite{series2012methodology}, three graduate students majoring in biomedical engineering are recruited to participate in subjective evaluation experiments. Prior to annotation, the participants undergo training using the scoring guidelines described in Section~\ref{sec:cas3k_rules}. After that, they are required to annotate $200$ test samples. To assess the reliability of their evaluations, two correlation metrics are calculated: the Pearson linear correlation coefficient (PLCC) and the Spearman rank correlation coefficient (SRCC). The formal annotation phase begins only after the inter-rater correlations exceeds a threshold of $0.7$.

During the formal annotation process, participants use a custom-designed scoring interface. As shown in Fig.~\ref{fig1} (a), each session displays three images simultaneously on a $27$-inch monitor with a resolution of $1920\times 1080$. Participants rate each image based on the three predefined quality metrics. To mitigate fatigue and maintain annotation quality, they are suggested to take breaks every $30$ minutes. In total, $3,565 \times 3\times 3 =32,085$ subject ratings are collected, where $3,565$, $3$, and $3$ represent the number of synthetic images, evaluation metrics, and subjects.

\subsection{Subjective Data Processing}
First, the outlier rejection procedure recommended by ITU-R BT.$500$~\cite{series2012methodology} are adopted to eliminate unreliable annotations. No outliers were detected and excluded in our case. To mitigate inter-observer variability in perceived image quality, per-subject score normalization is performed:
\begin{align}
    Z_{ij} = \frac{S_{ij}-\mu_{i}}{\sigma_{i}}
\end{align}
where $S_{ij}$ denotes the raw score assigned by the $i$-th subject to the $j$-th image, $\mu_{i}$ is the mean of scores and $\sigma_{i}$ is the standard deviation of scores from the $i$-th subject. Next, $Z_{ij}$ is linearly rescaled to $\left[0, 100\right]$ to obtain $\hat{Z}_{ij}$. Finally, the mean opinion score (MOS) is calculated following Eq.~(\ref{eq:1}), and the distributions of MOS are shown in Fig.~\ref{fig_mos}.
\begin{align}
    MOS_{j} = \frac{1}{N}\sum_{i=1}^N\hat{Z}_{ij} \label{eq:1}
\end{align}
where $N$ is the number of subjects.
\begin{figure}
\centering
\centerline{\includegraphics{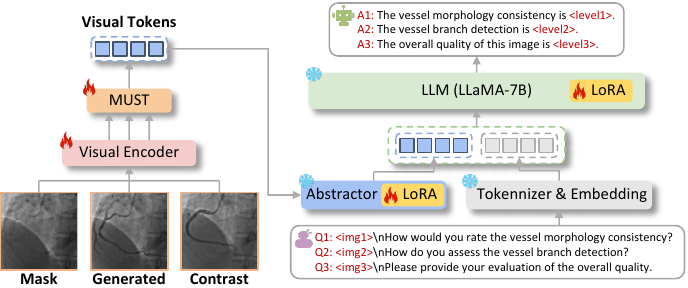}}
\caption{Overall framework of CAS-IQA.}
\label{fig2}
\end{figure}

\section{CAS-IQA Model} \label{sec:casiqa}
\subsection{Overall Framework}
As illustrated in Fig.~\ref{fig2}, CAS-IQA takes three images as input: a generated angiography (\textbf{Generated}), a non-contrast image (\textbf{Mask}), and a real angiography (\textbf{Contrast}). These images are first processed by a vision encoder to extract corresponding visual tokens. In parallel, manually crafted text prompts are tokenized and embedded into textual tokens. The proposed \underline{M}ulti-path feat\underline{U}re fu\underline{S}ion and rou\underline{T}ing (MUST) module then fuses the visual tokens from the three input images and routes them into three specialized branches, each tailored to evaluate a specific quality metric. The fused visual tokens are subsequently processed by a vision abstractor, which compresses them using learnable tokens. Finally, the visual and textual tokens are jointly fed into a large language model (LLM) to generate quality assessment predictions.

\subsection{Multi-Path Feature Fusion and Routing Module}
Experienced radiologists tend to focus on different regions of \textbf{Mask}, \textbf{Generated}, and \textbf{Contrast} depending on the quality metric being assessed. For VMC and VBD, attention is primarily directed toward vessels in \textbf{Generated} and \textbf{Contrast}. Specifically, VMC emphasizes the global structural alignment of vessels, whereas VBD requires attention to finer details, such as vessel bifurcations. In contrast, OQ involves a comprehensive evaluation of the overall consistency across all three images.
\begin{figure}
\centering
\centerline{\includegraphics{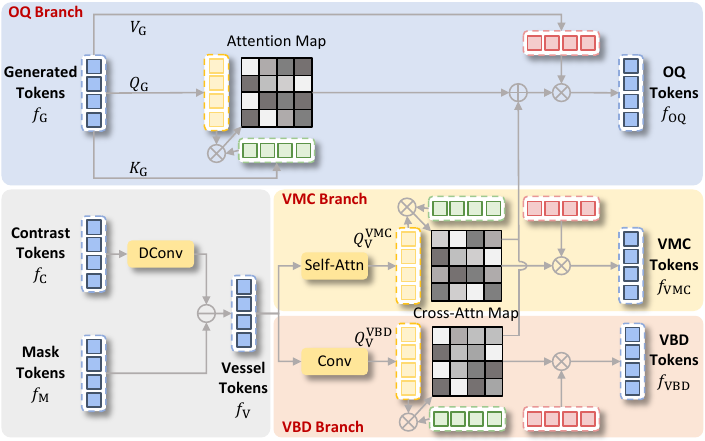}}
\caption{Details of the MUST module.}
\label{fig3}
\end{figure}
Motivated by these observations, a \underline{M}ulti-path feat\underline{U}re fu\underline{S}ion and rou\underline{T}ing (MUST) module is proposed, which enables fine-grained fusion and adaptive routing of visual tokens for each metric. The architecture of MUST is illustrated in Fig.~\ref{fig3}. For the VMC and VBD branches, we first extract vessel tokens by subtracting \textbf{Contrast} and \textbf{Mask} tokens. Since \textbf{Contrast} and \textbf{Mask} are not temporally aligned, deformable convolution is employed to align \textbf{Contrast} tokens before subtraction:
\begin{align}
    f_{\rm V} = {\rm DConv}\left(f_{\rm C}\right) - f_{\rm M}
\end{align}
where $f_{\rm V}$ represents vessel tokens. $f_{\rm C}$ and $f_{\rm M}$ denote \textbf{Contrast} and \textbf{Mask} tokens, respectively. ${\rm DConv}\left(\cdot\right)$ means deformable convolution.

\noindent \textbf{VMC Branch.} To evaluate global structural consistency, a self-attention mechanism~\cite{vaswani2017attention} is applied to vessel tokens $f_{\rm V}$ to obtain global vessel representations. These representations serve as queries in a cross-attention mechanism with \textbf{Generated} tokens, producing VMC-specific fused tokens $f_{\rm VMC}$.
\begin{align}\label{eqn-3}
A_{\rm VMC} = {\rm Softmax}\left(\frac{Q_{\rm V}^{\rm VMC}K_{\rm G}^{T}}{\sqrt{d}}\right),
f_{\rm VMC} = A_{\rm VMC}V_{\rm G}
\end{align}
where $A_{\rm VMC}$ is cross-attention maps. $Q_{\rm V}^{\rm VMC}=W_{\rm Q}^{\rm VMC}\text{Self-Attn}\left(f_{\rm V}\right)$ indicate VMC queries. $W_{\rm Q}^{\rm VMC}$ and $\text{Self-Attn}(\cdot)$ represent learnable projection matrices and self-attention, respectively. Keys $K_{\rm G}$ and values $V_{\rm G}$ are from \textbf{Generated} tokens.

\noindent \textbf{VBD Branch.} For VBD, which emphasizes local branching structures, fine-grained tokens are extracted from $f_{\rm V}$ using a convolutional block. These local tokens are then used to compute cross-attention with \textbf{Generated} tokens, producing VBD tokens $f_{\rm VBD}$:
\begin{align}\label{eqn-4}
A_{\rm VBD} = {\rm Softmax}\left(\frac{Q_{\rm V}^{\rm VBD}K_{\rm G}^{T}}{\sqrt{d}}\right), 
f_{\rm VBD} = A_{\rm VBD}V_{\rm G}
\end{align}
where VBD queries $Q_{\rm V}^{\rm VBD}=W_{\rm Q}^{\rm VBD}\text{Conv}\left(f_{\rm V}\right)$. $\text{Conv}\left(\cdot\right)$ is the convolutional block.

\noindent \textbf{OQ Branch.} Finally, to evaluate OQ, we begin by computing a self-attention map from \textbf{Generated} tokens to capture overall perceptual quality. This is then fused with the attention maps from the VMC and VBD branches using a set of learnable scalar weights $\alpha$, $\beta$ and $\gamma$, constrained such that $\alpha+\beta+\gamma=1$. The final attention map is applied to \textbf{Generated} values to produce OQ tokens:
\begin{align}\label{eqn-5}
f_{\rm OQ} = \left\{\alpha\left[{\rm Softmax}\left(\frac{Q_{\rm G}K_{\rm G}^{T}}{\sqrt{d}}\right)\right] + \beta A_{\rm VMC} + \gamma A_{\rm VBD} \right\}V_{\rm G}
\end{align}
\subsection{Mapping Between Levels and Scores}
\noindent \textbf{Continuous Scores to Discrete Levels.} Considering that humans tend to describe image quality in categorical levels rather than precise numerical scores~\cite{wu2024qalign}, a five-level rating scheme is adopted. The continuous score range of $0\sim 100$ is evenly partitioned into five intervals corresponding to the levels: \{bad, poor, fair, good, excellent\}. Each continuous score is mapped to its corresponding discrete level based on these intervals. These discrete labels are then used to construct the instruction–response pairs described in Section~\ref{casiqa_it} for supervised training.

\noindent \textbf{Score Estimation During Inference.} As illustrated in Fig.~\ref{fig1} (b), a softmax function is applied over the logits associated with the five predefined level tokens to obtain a probability distribution. For each quality level, a weight equal to the midpoint of its associated score interval is assigned. The final predicted quality score is computed as the weighted sum of these probabilities.

\subsection{Instruction Tuning}~\label{casiqa_it}
Instruction tuning has proven effective in unlocking the potential of VLMs for downstream tasks, and the design of high-quality user prompts is crucial for supervised fine-tuning~\cite{liu2023visual}. To improve model performance on the X-ray angiography IQA task, we design a series of prompts framed as multi-turn question–answer interactions.
\vspace{-0.8\baselineskip}
\begin{center}
\fcolorbox{black}{gray!10}{\parbox{1\linewidth}{
\noindent \textbf{\textit{Vessel Morphology Consistency:}}\\
\texttt{\#}\textbf{User:} {\tt<img1>}As an experienced interventional radiologist, how would you rate the vessel morphology consistency of this image?\\
\texttt{\#}\textbf{Assistant:} The vessel morphology consistency of this image is {\tt<level1>}.

\noindent \textbf{\textit{Vessel Branch Detection:}} \\
\texttt{\#}\textbf{User:} {\tt<img2>}From your perspective as an interventional radiologist, how do you assess the vessel branch detection in this image?\\
\texttt{\#}\textbf{Assistant:} The vessel branch detection of this image is {\tt<level2>}.

\noindent \textbf{\textit{Overall Quality:}} \\
\texttt{\#}\textbf{User:} {\tt<img3>}With your experience in interventional radiology, please provide your evaluation of the overall quality of this image.\\
\texttt{\#}\textbf{Assistant:} The overall quality of this image is {\tt<level3>}.
}}
\end{center}
\begin{table}[htbp]
\caption{Comparisons with SOTA methods on the CAS-3K dataset. VMC, VBD, and OQ denote vessel morphology consistency, vessel branch detection, and overall quality, respectively. The best and second results are highlighted in \sethlcolor{red!20}\hl{red} and \sethlcolor{blue!20}\hl{blue}, respectively.} \label{tab1}
\centering
\begin{tabularx}{\textwidth}{cc|*{8}{>{\centering\arraybackslash}X}} 
\toprule
\multicolumn{2}{c|}{\multirow{2}{*}{Category}} & 
\multicolumn{2}{c|}{\multirow{2}{*}{Methods}} & 
\multicolumn{2}{c}{VMC} & 
\multicolumn{2}{c}{VBD} & 
\multicolumn{2}{c}{OQ} \\
\cmidrule(lr){5-6} \cmidrule(lr){7-8} \cmidrule(lr){9-10}
\multicolumn{2}{c|}{} & 
\multicolumn{2}{c|}{} & 
PLCC & SRCC & PLCC & SRCC & PLCC & SRCC \\
\midrule
\multicolumn{2}{c|}{\multirow{2}{*}{Handcrafted}} &
\multicolumn{2}{c|}{NIQE~\cite{mittal2012making} {\tiny\color{gray}[SPL'12]}} & $0.0520$ & $0.0438$ & $0.0289$ & $0.0256$ & $0.0337$ & $0.0202$ \\
\multicolumn{2}{c|}{} &
\multicolumn{2}{c|}{BRISQUE~\cite{mittal2012no} {\tiny\color{gray}[TIP'12]}} & $0.1876$ & $0.1901$ & $0.2359$ & $0.2391$ & $0.2224$ & $0.2139$ \\
\midrule
\multicolumn{2}{c|}{\multirow{6}{*}{DNN-based}} &
\multicolumn{2}{c|}{DBCNN~\cite{Zhang_2020} {\tiny\color{gray}[TCSVT'20]}} & $0.4990$ & $0.4853$ & $0.5856$ & $0.5647$ & $0.4745$ & $0.4666$ \\
\multicolumn{2}{c|}{} &
\multicolumn{2}{c|}{HyperIQA~\cite{su2020blindly} {\tiny\color{gray}[CVPR'20]}} & $0.6657$ & $0.6356$ & $0.5942$ & $0.5873$ & $0.6339$ & $0.6216 $\\
\multicolumn{2}{c|}{} &
\multicolumn{2}{c|}{AHIQ~\cite{lao2022attentions} {\tiny\color{gray}[CVPR'22]}} & $0.6251$ & $0.6434$ & $0.6050$ & $0.6010$ & $0.6654$ & $0.6621$ \\
\multicolumn{2}{c|}{} &
\multicolumn{2}{c|}{ManIQA~\cite{yang2022maniqa} {\tiny\color{gray}[CVPR'22]}} & $0.6823$ & \cellcolor{blue!20}$0.6759$ & $0.6089$ & $0.6060$ & $0.6778$ & $0.6689$ \\
\multicolumn{2}{c|}{} &
\multicolumn{2}{c|}{QCN~\cite{shin2024blind} {\tiny\color{gray}[CVPR'24]}} & $0.6750$ & $0.6651$ & $0.6308$ & $0.6235$ & $0.6796$ & \cellcolor{blue!20}$0.6765$ \\
\multicolumn{2}{c|}{} &
\multicolumn{2}{c|}{LoDa~\cite{xu2024boosting} {\tiny\color{gray}[CVPR'24]}} & $0.6859$ & $0.6664$ & $0.6353$ & $0.6279$ & \cellcolor{blue!20}$0.6899$ & $0.6737$ \\
\midrule
\multicolumn{2}{c|}{\multirow{3}{*}{VLM-based}} &
\multicolumn{2}{c|}{Q-Align~\cite{wu2024qalign} {\tiny\color{gray}[ICLR'24]}} & $0.6030$ & $0.5907$ & $0.5683$ & $0.5651$ & $0.6096$ & $0.5999$ \\
\multicolumn{2}{c|}{} &
\multicolumn{2}{c|}{MA-AGIQA~\cite{wang2024large} {\tiny\color{gray}[MM'24]}} & \cellcolor{blue!20}$0.6892$ & $0.6625$ & \cellcolor{blue!20}$0.6507$ & \cellcolor{blue!20}$0.6314$ & $0.6696$ & $0.6598$ \\
\multicolumn{2}{c|}{} 
& \multicolumn{2}{c|}{\textbf{CAS-IQA (Ours)}}
& \cellcolor{red!20} $0.6925$
& \cellcolor{red!20} $0.6855$
& \cellcolor{red!20} $0.6639$
& \cellcolor{red!20} $0.6605$
& \cellcolor{red!20} $0.6985$
& \cellcolor{red!20} $0.6986$ \\
\bottomrule
\end{tabularx}
\end{table}

Note that the visual tokens {\tt <img1>}, {\tt <img2>}, and {\tt <img3>} correspond to the outputs of the VMC, VBD, and OQ branches, respectively. The predicted discrete quality levels for the VMC, VBD, and OQ metrics are represented as {\tt <level1>}, {\tt <level2>}, and {\tt <level3>}, respectively. Following Q-Align~\cite{wu2024qalign}, the user queries are randomly chosen from a group of paraphrases as augmentation.

\section{Experiments} \label{sec:exp}
\subsection{Implementation Details}
\textbf{Model Configurations.} Our CAS-IQA is built upon mPLUG-Owl2~\cite{ye2024mplug}, consisting of a pretrained CLIP image encoder (ViT-L), a visual abstractor (Q-Former), and a LLM (LLaMA-2-7B). Due to computational constraints and the limited size of CAS-3K, both the visual abstractor and the LLM are fine-tuned using low-rank adaptation (LoRA)~\cite{hu2022lora}. The LoRA configuration uses a rank of $128$ and an alpha value of $256$.

\noindent \textbf{Training Details.} CAS-3K is partitioned into a training set of $2,850$ images and a test set of $715$ images. Training is performed for $5$ epochs on two NVIDIA A6000 GPUs with a batch size of $16$. CAS-IQA is optimized using the AdamW optimizer with a learning rate of $2e^{-5}$, scheduled via cosine annealing.
\begin{figure}
\centering
\centerline{\includegraphics{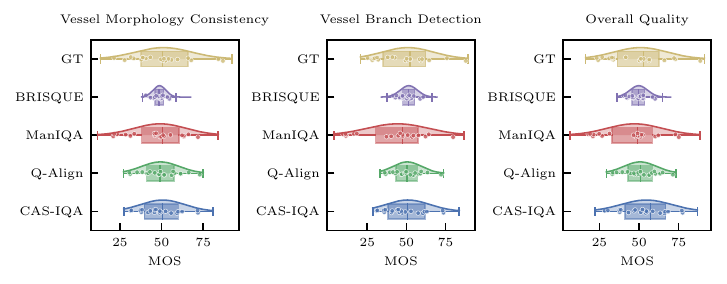}}
\caption{MOS distributions predicted by different methods.}
\label{box_plot}
\end{figure}

\noindent \textbf{Evaluation Metrics.} In line with prior work~\cite{wu2024qalign,you2025teaching}, we evaluate model performance using the Pearson linear correlation coefficient (PLCC) and the Spearman rank-order correlation coefficient (SRCC). Higher PLCC and SRCC values indicate stronger alignment between the predicted quality scores and human subjective judgments.

\subsection{Comparisons with SOTA Methods}
\textbf{Quantitative Results.} To demonstrate the effectiveness of CAS-IQA, we compare it against a comprehensive set of baselines, including two handcrafted IQA methods, six DNN-based approaches, and two VLM-based methods. Since these baseline models are designed to predict a single quality score, each of them are separately trained for the three metrics. Quantitative results is summarized in Table~\ref{tab1}. Our main observations are as follows: \textbf{1)} Handcrafted methods (\textit{i.e.,} NIQE~\cite{mittal2012making} and BRISQUE~\cite{mittal2012no}) perform poorly on the CAS-3K dataset. These methods rely on handcrafted statistical features optimized for natural image statistics. The assumptions about luminance distribution, texture, and contrast do not hold in the domain of X-ray angiographies. Such methods lack the representational capacity to capture the anatomical structures and modality-specific patterns inherent in synthetic angiographies. Consequently, the poor performance of these methods is expected; \textbf{2)} Our CAS-IQA achieves superior results compared to all baseline methods. Notably, it achieves PLCC improvements of $0.5\%$, $2.0\%$, and $4.1\%$ for VMC, VBD, and OQ, respectively, over the second-best method, MA-AGIQA~\cite{wang2024large}. These results are consistent with our design rationale. Baseline models typically rely solely on \textbf{Generated} and therefore lack critical auxiliary information. In contrast, CAS-IQA integrates visual cues from multiple inputs and adaptively routes metric-specific visual cues to dedicated branches. This enables CAS-IQA to perform more accurate quality assessment.

\noindent \textbf{Qualitative Analysis.} 
Fig.~\ref{box_plot} illustrates the distributions of MOS predicted by four representative methods across the three evaluation metrics. We present a detailed analysis as follows: \textbf{1)} The MOSs predicted by handcrafted methods tends to concentrate within a narrow range. This is largely attributed to the inherent domain gap between natural images. Unlike natural images, which typically contain rich textures and diverse color distributions, X-ray angiographies are structurally homogeneous and feature-sparse. As a result, these approaches fail to capture meaningful variations in image quality, mistakenly interpreting all inputs as perceptually similar;\begin{table}[htbp]
\caption{
Quantitative results of ablation study. The best results are highlighted in \textbf{bold}.
} \label{tab2}
\centering
\begin{tabularx}{\textwidth}{c|*{6}{>{\centering\arraybackslash}X}}
\toprule
\multirow{2}{*}{MUST} & 
\multicolumn{2}{c}{VMC} &
\multicolumn{2}{c}{VBD} & 
\multicolumn{2}{c}{OQ} \\
\cmidrule(lr){2-3} \cmidrule(lr){4-5} \cmidrule(lr){6-7}
& PLCC & SRCC & PLCC & SRCC & PLCC & SRCC \\
\midrule
\ding{55} & $0.6517$ & $0.6636$ & $0.6451$ & $0.6491$ & $0.6830$ & $0.6750$ \\
\ding{51} & $\textbf{0.6925}$ & $\textbf{0.6855}$ & $\textbf{0.6639}$ & $\textbf{0.6605}$ & $\textbf{0.6985}$ & $\textbf{0.6986}$ \\
\bottomrule
\end{tabularx}
\end{table} \textbf{2)} DNN-based methods, such as ManIQA~\cite{yang2022maniqa}, exhibit similar distribution patterns across three metrics, primarily due to their reliance solely on human-annotated supervision without the benefit of prompt-driven guidance as seen in VLM-based approaches; \textbf{3)} Previous VLM-based methods, such as Q-Align~\cite{wu2024qalign}, give score predictions only based on \textbf{Generated}, without incorporating auxiliary structural cues in \textbf{Mask} and \textbf{Contrast}. This lack of contextual information limits their semantic grounding and impairs their ability to assess image quality in a clinically meaningful way. In contrast, CAS-IQA, equipped with MUST module, effectively fuses and routes multi-source image features, enabling more accurate and metric-specific quality assessment.

\subsection{Ablation Study}
\textbf{Quantitative Results.} The effectiveness of the proposed MUST module is evaluated in Table~\ref{tab2}. In the baseline configuration (w/o MUST), a simplified setup is used in which a single prompt processes visual tokens from \textbf{Mask}, \textbf{Generated}, and \textbf{Contrast} without specialized fusion or routing. 
\vspace{-0.8\baselineskip}
\begin{center}
\fcolorbox{black}{gray!10}{\parbox{1\linewidth}{
\noindent\texttt{\#}\textbf{User:} {\tt<imgG>}As an experienced interventional radiologist, how would you rate the quality of this image? Contrast-free image:{\tt<imgM>}, Contrast imgae:{\tt<imgC>}\\
\texttt{\#}\textbf{Assistant:} The vessel morphology consistency of this image is {\tt<level1>}. The vessel branch detection rate of this image is {\tt<level2>}. The overall quality of this image is {\tt<level3>}.
}}
\end{center}

\begin{figure}
\centering
\centerline{\includegraphics{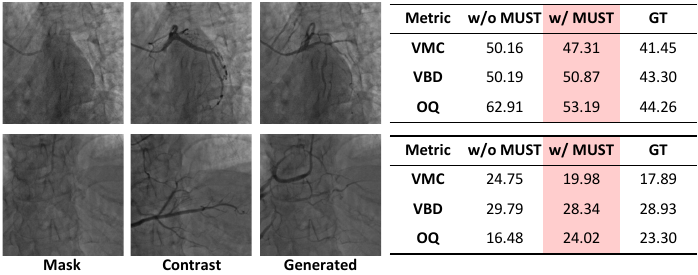}}
\caption{Two examples of CAS-IQA (w/ and w/o MUST).}
\label{ablation_vis}
\end{figure}

As shown in the results, incorporating MUST leads to consistent improvements across all three metrics. These improvements are primarily attributed to two key mechanisms within the MUST module: \textbf{1) Multi-Source Visual Fusion:} By explicitly fusing visual tokens from multiple images, MUST enables context-aware feature representation. This is particularly beneficial for metrics like OQ, which require a comprehensive assessment of visual coherence across inputs; \textbf{2) Metric-Specific Routing:} MUST introduces dedicated evaluation branches, allowing each metric to receive features aligned with its semantic requirements. For example, global vessel morphology for VMC and localized bifurcations for VBD. This targeted routing enhances metric-specific prediction accuracy. In addition to performance improvements, we also observed that CAS-IQA equipped with the MUST module achieves optimal performance within only 5 training epochs, whereas the baseline configuration (w/o MUST) requires around 15 epochs to converge. This acceleration in convergence can be attributed to several factors: First, multi-source fusion enables early-stage feature representations to be more discriminative and semantically informative, reducing the learning burden. Second, the metric-specific routing mechanism introduces task-specific inductive bias, allowing the model to optimize each evaluation metric independently and efficiently.

\noindent \textbf{Qualitative Analysis.} Fig.~\ref{ablation_vis} presents the quality score predictions of CAS-IQA (w/ and w/o MUST) on two representative examples. In both examples, integrating the MUST module enables CAS-IQA to produce predictions that are consistently closer to human subjective judgments (\textit{i.e.,} GT). However, we observe that the improvements brought by MUST in VBD are less pronounced compared to those in VMC and OQ. One possible explanation is that \textbf{Generated} often contain isolated or anatomically implausible vessel branches. Such artifacts are visually distinct and can be detected directly without the need for additional contextual information. This indicates that VBD, being more anomaly-sensitive, may rely more on visual saliency within \textbf{Generated} than on broader contextual consistency. In contrast, VMC and OQ require a holistic understanding of vessel morphology and global image coherence. Accurate assessment of these metrics benefits from integrating structural information across multiple images. The design of the MUST module facilitates this process by enabling multi-source feature integration and metric-specific representation learning.

\section{Conclusion} \label{sec:conclusion}
This paper focuses on image quality assessment (IQA) for synthetic X-ray angiographies. To enable clinically meaningful evaluation, a curated dataset, CAS-3K, is constructed, and a unified framework, CAS-IQA, is proposed. By leveraging the powerful vision-language reasoning capabilities of VLMs, CAS-IQA can achieve strong alignment with human subjective judgments. In addition, the proposed MUST module enhances performance by adaptively fusing image tokens and routing them to metric-specific branches. Experimental results indicate that CAS-IQA achieves the state-of-the-art performance. This work unlocks the potential of VLMs in accurate IQA for synthetic X-ray angiographies. Future work will focus on improving the interpretability of CAS-IQA to facilitate its deployment in real-world clinical applications.

\section*{Acknowledgements}
This work was supported in part by the National Key Research and Development Program of China under Grant 2023YFC2415100, in part by the National Natural Science Foundation of China under Grant 62222316, Grant 62373351, Grant 82327801, Grant 62073325, Grant 62303463, in part by the Chinese Academy of Sciences Project for Young Scientists in Basic Research under Grant No. YSBR-104, in part by the Beijing Natural Science Foundation under Grant F252068, Grant 4254107, in part by China Postdoctoral Science Foundation under Grant 2024M763535 and in part by CAMS Innovation Fund for Medical Sciences (ClFMS) under Grant 2023-I2M-C\&T-B-017.

\bibliographystyle{splncs04}
\bibliography{ref}

\end{document}